\documentclass[letterpaper, 10 pt, conference]{ieeeconf}  

\IEEEoverridecommandlockouts                              

\usepackage{graphicx}
\usepackage{amsmath} 
\usepackage{amssymb}  

\title{\LARGE \bf
Classification based Grasp Detection using Spatial Transformer Network
}

\author{Dongwon Park$^{1}$ and Se Young Chun$^{1,\dagger}$
\thanks{*This work was supported by the Technology Innovation Program or 
Industrial Strategic Technology Development Program 
(10077533, Development of robotic manipulation algorithm for grasping/assembling with 
the machine learning using visual and tactile sensing information) funded 
by the Ministry of Trade, Industry \& Energy (MOTIE, Korea).}
\thanks{$^{1}$Dongwon Park and Se Young Chun are with Department of Electrical Engineering, 
Ulsan National Institute of Science and Technology (UNIST), Ulsan, Republic of Korea.
       $^\dagger$Email: {\tt\small sychun@unist.ac.kr}}%
}

\begin{document}

\maketitle
\thispagestyle{empty}
\pagestyle{empty}

\begin{abstract}

Robotic grasp detection task is still challenging, particularly for novel objects.
With the recent advance of deep learning, there have been several works on
detecting robotic grasp using neural networks.
Typically, regression based grasp detection methods have outperformed 
classification based detection methods in computation complexity with excellent accuracy.
However, classification based robotic grasp detection still seems to have merits
such as intermediate step observability and straightforward back propagation routine for end-to-end training.
In this work, we propose a novel classification based robotic grasp detection method with
multiple-stage spatial transformer networks (STN).
Our proposed method was able to achieve 
state-of-the-art performance in accuracy with real-time computation.
Additionally, unlike other regression based grasp detection methods, 
our proposed method allows partial observation for intermediate results 
such as grasp location and orientation for a number of grasp configuration candidates.

\end{abstract}

\section{INTRODUCTION}

Robotic grasping of novel objects is still a challenging problem. 
It requires to perform robotic grasp detection, trajectory planning and execution.
Detecting robotic grasp from imaging sensors is a crucial step for successful grasping.
There have been numerous works on robotic grasp detection (or synthesis).
In large, grasp synthesis is divided into
analytical or empirical (or data-driven) methods~\cite{Sahbani:2012kp}
for known, familiar or novel objects~\cite{Bohg:2014ef}. 

Machine learning approaches for robotic grasp detection have utilized data to learn
discriminative features for a suitable grasp configuration and to yield 
excellent performance on generating grasp locations~\cite{Saxena:2008doa,Bohg:2010es}. 
Since deep learning has been successful in computer vision applications such as 
image classification~\cite{Krizhevsky:2012wl,He:2016ib} and object detection~\cite{Ren:2015ug,Redmon:2017gn}, 
these powerful tools have applied to 
robotic grasp detection of location and orientation~\cite{Lenz:2015ih,Redmon:2015eq}.

There are two types of machine learning approaches in robotic grasp detection of location and orientation.
One is classification based approach that trains classifiers to discriminate graspable points and orientations for 
local image patches~\cite{Saxena:2008doa,Bohg:2010es,Lenz:2015ih,Jiang:2011ja}. 
In general, sliding window approach generates numerous candidates for robotic grasp configurations with different location,
orientation and scale (size of gripper). Then, these image patches are fed into machine learning (or deep learning) 
classifiers to yield the scores of graspability (the higher score, the better graspability). 
Finally, one candidate image patch with the highest score will be selected. The location, orientation 
and size of that image patch will be the final grasp detection result. 
Unfortunately, this approach is in general slow due to brute-force sliding window.
Note that classifier based robotic grasp detection is similar to
grasp detection methods using 3D grasp simulators~\cite{Miller:2003ev,Leon:2010jf}.

\begin{figure}[!t]
      \centering
      \includegraphics[width=0.6\linewidth]{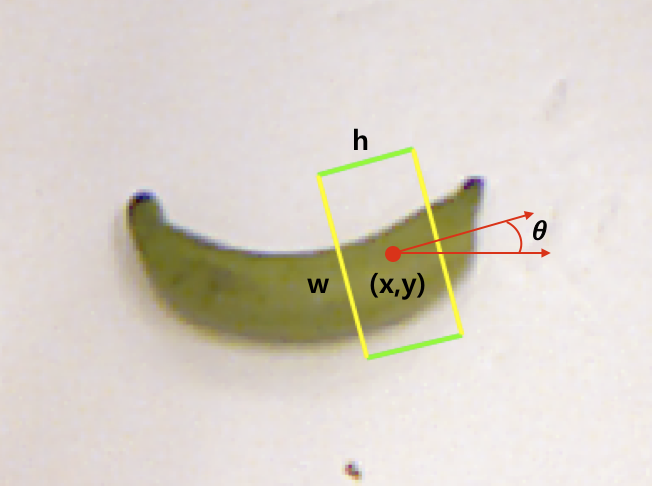}
      \caption{An example of a robotic grasp detection with five-dimensional grasp representation for a banana.
	Green lines are two plates of a gripper whose size is $h$, yellow lines are the distance between two
	plate grippers for grasping, red point is the center location of grasp rectangle $(x, y)$, and red angle
	$\theta$ is the orientation of the grasp rectangle.}
      \label{fig:goodgrasp}
      	\vskip -0.2in
   \end{figure}

The other is regression based approach that trains a model (parametric or non-parametric, 
neural network or probability distribution) to yield robotic grasp detection parameters 
for location and orientation directly~\cite{Redmon:2015eq,Kumra:2017ko,Asif:2017bv}.
Typical five-dimensional robotic grasp parameters are shown in Fig.~\ref{fig:goodgrasp}, 
where the width ($w$), height ($h$), location ($x, y$) and  and orientation ($\theta$) of a
grasp rectangle~\cite{Jiang:2011ja,Lenz:2015ih}. Note that with additional depth and surface norm information,
these five parameters can be transformed into seven-dimensional robotic grasp representation~\cite{Jiang:2011ja}.
Since all grasp parameters are directly estimated from a single image (or a set of multi-modal images),
no sliding window is required.
Regression based robotic grasp detection methods are usually much faster
than classification based methods in term of computation~\cite{Redmon:2015eq}.
However, fair comparison between regression and classification based grasp detection
does not seem to be well investigated for different computation platforms.
   
   \begin{figure*}[!h]
      \centering
      \includegraphics[width=0.8\textwidth]{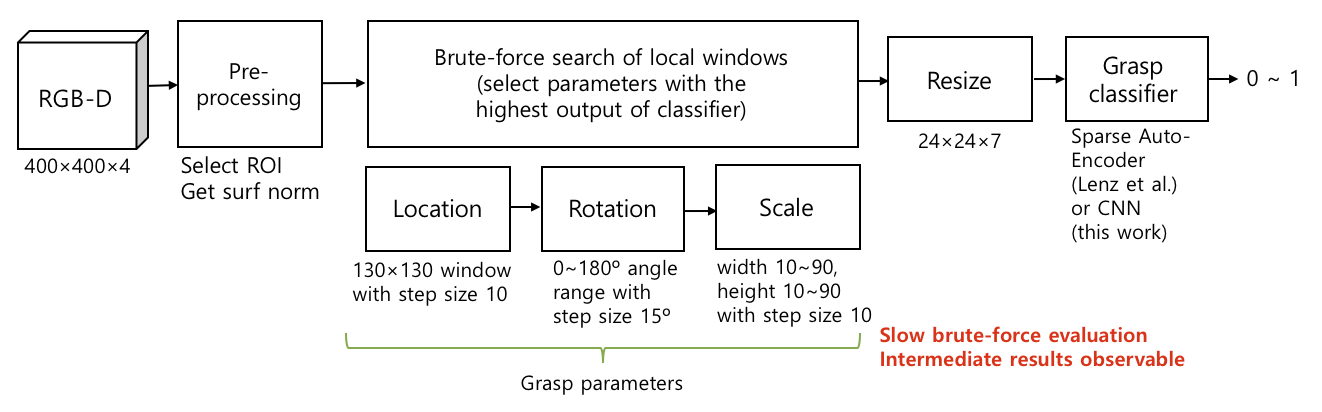}
      	\vskip -0.1in
      \caption{Classification based robotic grasp detection procedure~\cite{Lenz:2015ih}.
      Sliding window brute-force search of candidates consumes lots of computation time
      for different location, orientation, and scale. However, intermediate steps are clearly visible so that
      good or bad candidates can be observable.}
      \label{fig:lenz}
      	\vskip -0.1in
   \end{figure*}
   
In this paper, we propose a novel classification based robotic grasp detection method
using multiple-stage spatial transformer networks (STN). 
Unlike other black-box regression based grasp detection methods, 
multiple-stage STN of our proposed method allows partial observation of intermediate grasp results 
such as grasp location and orientation, for a number of candidates for grasp.
We will show that our proposed method achieves 
state-of-the-art performance in terms of both accuracy and computation complexity.
The contribution of this paper is as follows:
\begin{itemize}

\item A novel multiple-stage STN network is proposed using the original STN~\cite{Jaderberg:2015voa}
and the deep residual network (ResNet)~\cite{He:2016ib} instead of brute force sliding window. 
Intermediate grasp results are now partially observable in our proposed network.

\item A new classification based robotic grasp detection method using our multiple-stage STN 
is proposed for real-time detection with excellent accuracy. 
End-to-end training strategy was investigated for our proposed method with high resolution images.

\item Extensive comparison of our proposed method with other methods on the same platform was
performed with the same size of input image for fair comparison.

\end{itemize}

\section{RELATED WORK}

\subsection{Deep Learning based Object Detection}

There have been much research on object detection using deep learning. Object detection is simply to
identify the location and the class of an object (or objects).

There have been several classifier based object detection methods proposed such as
region-based convolutional neural network (R-CNN)~\cite{Girshick:2014jx}, 
fast R-CNN~\cite{Girshick:2015ib} and faster R-CNN~\cite{Ren:2015ug}. 
The original R-CNN uses a classifier for a local patch of an image with
sliding window on the image to identify where and what the object is~\cite{Girshick:2014jx}.
Due to time-consuming sliding window with heavy CNN operations, this method is known to be slow.
This sliding window approach is similar to the work of Lenz \textit{et al.} in robotic grasp detection problem~\cite{Lenz:2015ih}.
Fast R-CNN significantly reduced computation cost of R-CNN by having sliding window not on an image, but on a feature
space~\cite{Girshick:2015ib}. However, due to a large amount of object detection candidates, 
it was still not a real-time processing.
Faster R-CNN proposed region proposal network (RPN) to reduce the amount of object detection candidates so that it significantly
improved computation time~\cite{Ren:2015ug}. RPN generates candidate rectangles of object detection selectively
so that faster R-CNN maintains (or improves) detection performance while reduces computation.

Several regressor based object detection methods have also been proposed such as
you only look once (YOLO)~\cite{Redmon:2016gh}, 
single shot multibox detector (SSD)~\cite{Liu:2016bw} and
YOLO9000~\cite{Redmon:2017gn}. Instead of evaluating many object detection candidate windows,
these methods process an image only once to estimate object detection rectangles directly so that
fast computation is able to be achieved.
YOLO used a pre-trained AlexNet~\cite{Krizhevsky:2012wl} to estimate the location and class of
multiple objects~\cite{Redmon:2016gh}. SSD further developed regression based
object detection to incorporate intermediate
CNN features for object detection and improved accuracy and computation speed~\cite{Liu:2016bw}.
Recently, YOLO9000 extended the original YOLO significantly to classify 9000 classes of objects
with fast computation and high accuracy~\cite{Redmon:2017gn}.
These approaches are similar to the work of Redmon and Angelova~\cite{Redmon:2015eq} in robotic grasp detection.

      \begin{figure*}[!h]
      \centering
      \includegraphics[width=0.8\textwidth]{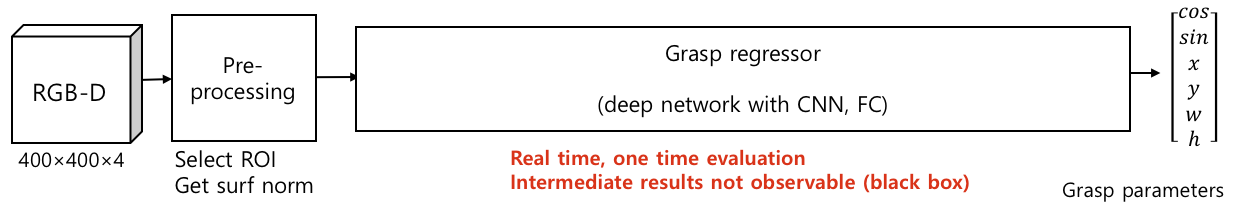}
      	\vskip -0.1in
      \caption{Regression based robotic grasp detection pipeline~\cite{Redmon:2015eq}.
      One shot evaluation for the whole image is possible to directly generate grasp configuration such as 
      location, orientation, and scale. However, intermediate steps are not observable.}
      \label{fig:redmon}
      \vskip -0.1in
   \end{figure*}
   
\subsection{Deep Learning based Robotic Grasp Detection}

Data-driven robotic grasp detection for novel object has been investigated extensively~\cite{Bohg:2014ef}.
Before deep learning, there have been some works on grasp location detection using machine learning
techniques~\cite{Saxena:2008doa}. Saxena \textit{et al.} proposed a machine learning method to rank 
the best graspable location for all candidate image patches from different locations.
Jiang \textit{et al.} proposed to use a five-dimensional robotic grasp representation 
as also shown in Fig.~\ref{fig:goodgrasp} and further improved a machine learning method to rank the best
graspable image patch whose representation includes orientation and gripper distance 
among all candidates~\cite{Jiang:2011ja}.

Since the advent of deep learning~\cite{LeCun:2015dt}, robotic grasp detection using deep learning has been
investigated for improved accuracy.
Lenz \textit{et al.} proposed a sparse auto-encoder (SAE) to train the network to rank the best
graspable image patch with multi-modal information (RGB color, depth, and calculated surface norm)
and to apply to robotic grasp detection using sliding window~\cite{Lenz:2015ih}.
However, due to time-consuming sliding window process, this method was slow (13.5 sec per image).
Moreover, this work was not further extended with simple modification using recent activation functions such
as ReLU (Rectifier Linear Unit), which could potentially improve performance significantly~\cite{Krizhevsky:2012wl}.
This method can be categorized into classification based grasp detection method.
This type of methods allows to observe intermediate steps of grasp detection by showing many candidates
with good or bad graspability.
Fig.~\ref{fig:lenz} illustrates an example of classification based robotic grasp detection pipeline.

Then, Redmon and Angelova proposed a real-time robotic grasp detection 
using modified AlexNet~\cite{Krizhevsky:2012wl,Redmon:2015eq}. 
A pre-trained AlexNet was modified to estimate robotic grasp parameters directly for local windows so that
no sliding windows is necessary. This approach significantly improved the performance of grasp detection
in accuracy and computation time over the work of Lenz \textit{et al.}.
To incorporate depth information without changing the network structure much, blue channel was used for depth.
No sliding window seems to contribute to the speed up of this method significantly and 
recent deep network seems to help achieving state-of-the-art accuracy.
This method can be categorized into regression based grasp detection method.
Unfortunately, this type of regression based methods do not allow 
to observe intermediate steps of grasp detection.
Fig.~\ref{fig:redmon} illustrates an example of regression based robotic grasp detection procedure.

Recently, Asif \textit{et al.} proposed a object recognition and grasp detection using
hierarchical cascaded forests using features extracted from deep learning~\cite{Asif:2017bv}. 
This work focused on improving accuracy of tasks, rather on improving computation time.
Wang \textit{et al.} proposed a real-time classification based grasp detection method using 
two-stage approach~\cite{Wang:2016cpa}. This method utilized a stacked SAE for classification, which
is similar to the work of Lenz \textit{et al.}, but with much more efficient grasp candidate generation.
This method utilized several prior information and pre-processing to reduce
the search space of grasp candidates 
such as object recognition result and the graspability of previously
evaluated image patches. It also reduced the number of grasp parameters to estimate such as
height ($h$) for known gripper and the orientation ($\theta$) 
that could be analytically calculated from surface norm.
This model does not support end-to-end learning for candidate estimation block.
Kumra and Kanan proposed a real-time regression based grasp detection method using
ResNet for multimodal information (RGB-D). Two pre-trained ResNet-50 networks were used to extract
features for RGB color image and for depth image, respectively~\cite{Kumra:2017ko}. 
Then, a neural network with 3 fully connected layers merged these feature vectors to yield
grasp configuration such as width, height, orientation and location.
End-to-end training optimized the whole network.
However, due to regression based approach, intermediate steps are not observable.

\section{PROPOSED METHOD}

\subsection{Multiple-Stage STN for Robotic Grasp Detection} 

Instead of slow sliding window for generating many grasp candidates, we propose a multi-stage STN for generating
a number of highly selective robotic grasp candidates. This approach seems similar to RPN in~\cite{Ren:2015ug}, but
RPN has never used in robotic grasp detection and can not deal with different orientation.
STN can explicitly encode spatial transformation including orientation~\cite{Jaderberg:2015voa}.
Our approach is also different from the work of Wang \textit{et al.}~\cite{Wang:2016cpa} that used prior information,
while our approach is fully data-driven and can support end-to-end training for fine tuning.

    \begin{figure}[b]
      \centering
      \vskip -0.1in
      \includegraphics[width=0.9\linewidth]{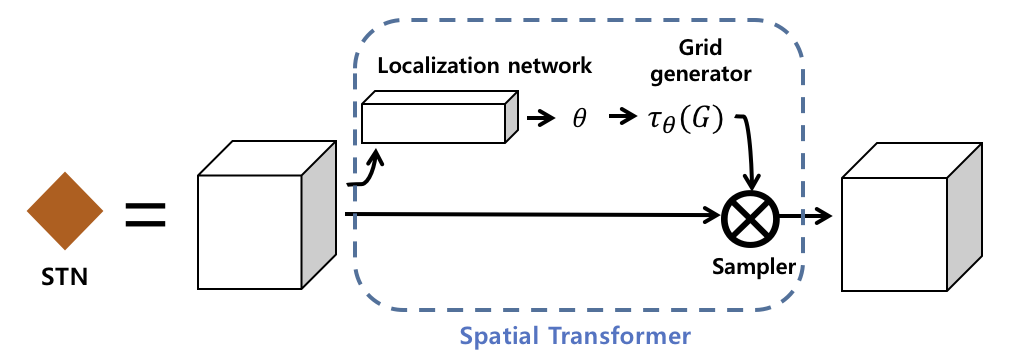}
      \caption{A typical STN structure proposed in~\cite{Jaderberg:2015voa}. Localization network generates
      transformation parameters. Then, STN transforms an input image or feature map using the output parameters.
      It is possible to back propagate this network since it is differentiable.}
      \label{fig:stn}
   \end{figure}

      \begin{figure*}[!h]
      \centering
      \includegraphics[width=0.8\textwidth]{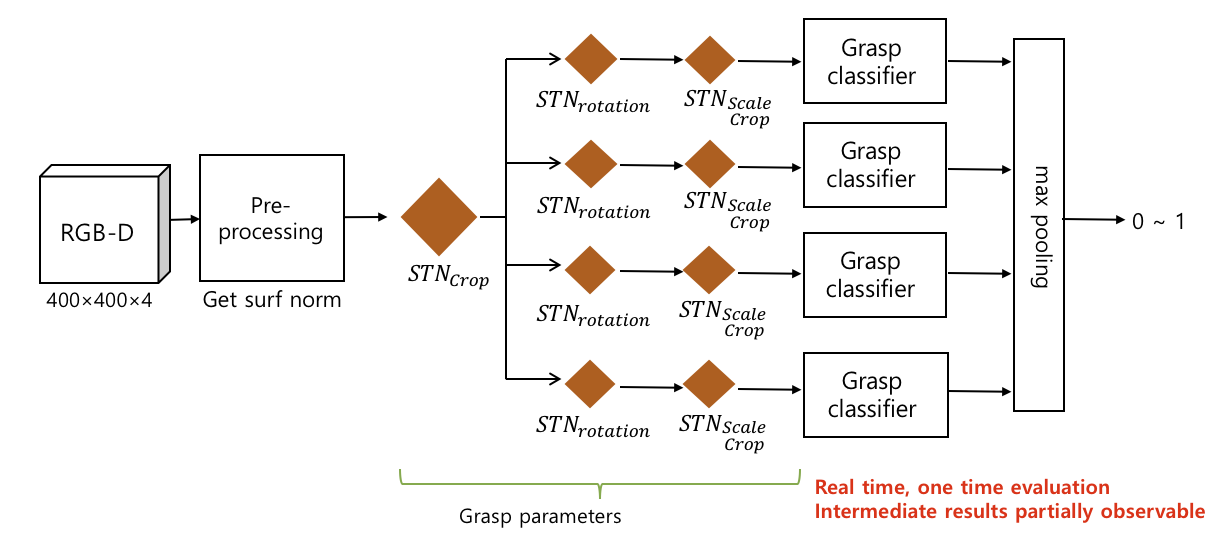}
      \vskip -0.1in
      \caption{Our proposed classification based robotic grasp detection pipeline.
      One shot evaluation for the whole image is possible to generate a number of potential grasp candidates
      so that it is fast as well as intermediate steps are partially observable.}
      \label{fig:multistn}
      \vskip -0.2in
   \end{figure*}
   
STN consists of localization network to generate transformation parameters, grid generators with the output
transformation parameters and sampler to generate warped images or feature maps 
as shown in Fig.~\ref{fig:stn}~\cite{Jaderberg:2015voa}.
We modified the original STN by constructing a new localization network using residual blocks~\cite{He:2016ib}.
STN can generate all necessary robotic grasp configuration in one network like regression based grasp detection method.
However, we adopt multiple-stage approach to generate a number of locations, then to estimate proper orientation for each 
location, and lastly to fine tune for locations and scale (width and height).
Note that this procedure seems similar to human grasp detection. 
Humans usually find possible grasp locations first, then estimate
orientation and scale information.
Our multiple-stage STN is illustrated in Fig.~\ref{fig:multistn}. Note that since
STN is differentiable, it is possible to implement back propagation for our proposed multiple-stage STN.

One of the potential advantages in our multiple-stage STN approach is that 
intermediate grasp detection steps are partially observable. A number of generated candidates
are observable at each stage of our proposed STN. This feature was practically helpful for us
to correct for potential errors in grasp detection during our experiment.
In addition, it is possible for our proposed multiple-stage STN to be trained end-to-end.
This is not only helpful for training robotic grasp detection with ground truth data with labels, but also potentially
useful for end-to-end learning from robotic control systems so that the whole network can be improved over
trial and error of robots~\cite{Levine:2016wg}.

\subsection{Classification based Grasp Detection using STN}

Our proposed grasp detection network using multiple-stage STN is illustrated in Fig.~\ref{fig:multistn}.
After pre-processing, $STN_{Crop}$ identifies a number of possible grasp locations and generates a set of locations $(x,y)$.
Then, for each location, $STN_{rotation}$ is applied to find appropriate orientation $\theta$ for graspable areas.
Lastly, $STN_{Scale, Crop}$ determines the width and height for gripper distance and size as well as 
additional locations $(\delta x, \delta y)$ for fine tuning.
Each cropped image patch is fed into grasp classifier to generate the score of graspability.
Finally, max pooling select the best graspable configuration among selective candidates.

For training this proposed network, first of all, each component was trained using ground truth data:
three STN networks as well as grasp classifier. Then, for end-to-end learning, we focus on
training the end block (grasp classifier) first, and then train the second end block ($STN_{Scale, Crop}$) while
fixing the previous block, and so on. This fine tuning step was able to improve accuracy further.
Note that unlike previous works using $224 \times 224$ input images, our proposed network
has relatively high resolution input images with $400 \times 400$, roughly 3 times more pixels than previous cases.

\section{EXPERIMENTAL RESULTS}

Most robotic grasp detection works compared their methods with other previous approaches
based on the reported results in papers. While the accuracy comparison may be reasonable,
computation complexity comparison may require more careful approach than that due to
many crucial factors such as input image size and used GPU spec.
In this work, we reproduced all the results of previous methods for comparison
on the same platform with the same size data.

\subsection{Dataset}

    \begin{figure}[!b]
    \vskip -0.1in
      \centering
      \includegraphics[width=0.65\linewidth]{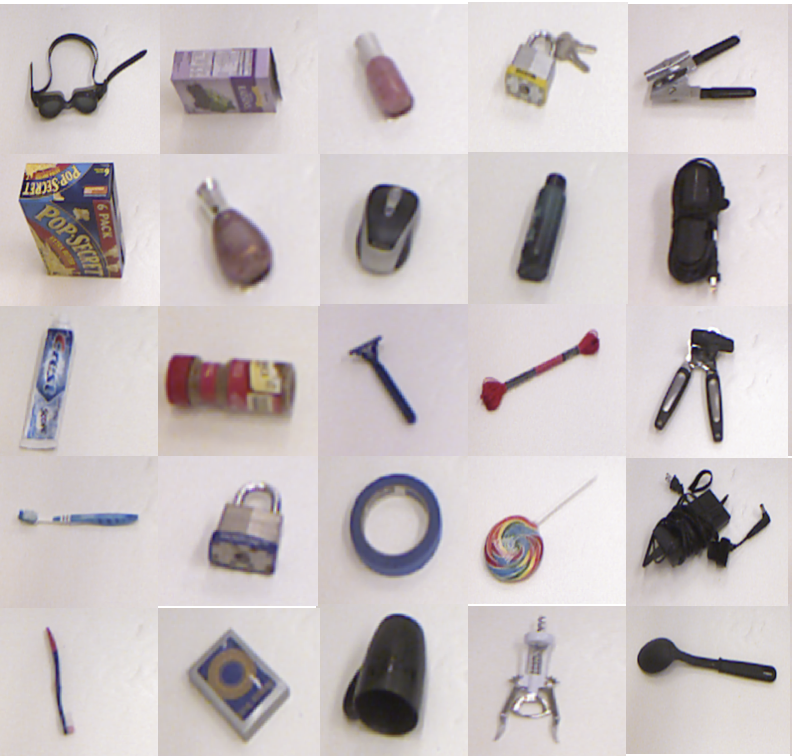}
      \caption{Examples of the Cornell grasp detection dataset~\cite{Lenz:2015ih}.}
      \label{fig:dataset}
   \end{figure}
      
We trained and tested our proposed classification based robotic grasp detection method using multiple-stage STN with 
the Cornell grasp detection dataset~\cite{Lenz:2015ih}.
This dataset contains 855 images (RGB color and depth) of 240 different objects with the ground truth labels of
a few graspable rectangles and a few not-graspable rectangles.
To train our grasp classifiers to assign low graspability score, we additionally generated a number of background
image patches with almost all white color. Note that we used cropped images with $400 \times 400$ instead of 
$224 \times 224$, which is relatively high resolution than the resolution previous works used. It was possible since
our proposed method does not require pre-training with massive dataset such as ImageNet.

\subsection{Training}

We split the Cornell grasp dataset into training set and test set (4:1) with image-wise spliting.
To train individual blocks of our proposed network as shown in Fig.~\ref{fig:stn}, 
ground truth labels were transformed into appropriate values for each STN block.
We set the output of $STN_{Crop}$ to be 8 so that 4 candidate locations can be generated for 
potential graspable areas. To generate the initial ground truth for this network,
we put the top right ground truth label as the first output of this network and put zero if 
there are not enough graspable labels.
Note that no data augmentation and no pre-training were used.

    \begin{figure}[!t]
      \centering
      \includegraphics[width=0.75\linewidth]{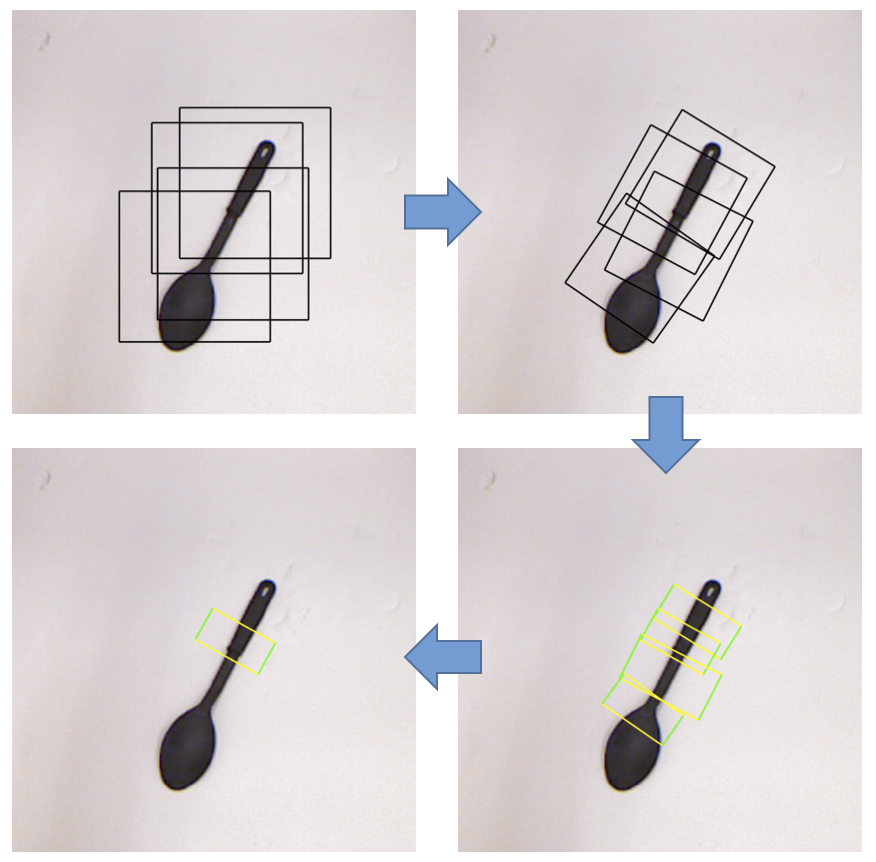}
      \caption{One result of our proposed grasp detection method. Top left figure is the output of 
      the first STN (4 locations) and top right is the output of the second STN with rotation.
      Then, the bottom right is the output of the last STN with scaling and fine tuned location.
      The bottom left is the best graspable configuration among 4 outputs of the last STN.}
      \label{fig:result1}
      \vskip -0.2in
   \end{figure}
   
\subsection{Evaluation}

The same metric for accuracy was used as in~\cite{Lenz:2015ih,Redmon:2015eq,Kumra:2017ko}.
When the difference between the output orientation and the ground truth orientation is less than 30$^o$,
the Jaccard index was measured between the output rectangle and the ground truth rectangle.
When the Jaccard index is more than 25\%, the output grasp configuration is considered as a good grasp 
and otherwise bad.
 
\subsection{Implementation}

The original Lenz \textit{et al.}'s result was reproduced 
using the MATLAB code provided by the authors~\cite{Lenz:2015ih} (called classification with SAE).
We optimized the original code further so that similar computation speed
was able to be achieved for larger input images with $400 \times 400$.
We also implemented the work of Lenz \textit{et al.} using Tensorflow and replaced SAE with 4-layer CNN 
(called classification with CNN). 
For direct regression based grasp detection method that is similar to~\cite{Redmon:2015eq}, 
we implemented it using Tensorflow based ResNet-32~\cite{He:2016ib}.
Unlike the original work using 3 channels (RG and Depth)~\cite{Redmon:2015eq},
we implemented the ResNet-32 with 7 channel input so that all multimodal information can be used 
(RGB color, depth, 3 channel surface norm). Thus, no pre-training was performed for this implementation.
Lastly, our proposed methods using single stage STN and multiple stage STN were implemented using Tensorflow.
All algorithms were tested on the platform with a single GPU (NVIDIA GeForce GTX 1080 Ti), 
a single CPU (Intel i7-7700K \@ 4.20GHz) and 32 GB memory.

    \begin{figure}[!t]
      \centering
      \includegraphics[width=0.75\linewidth]{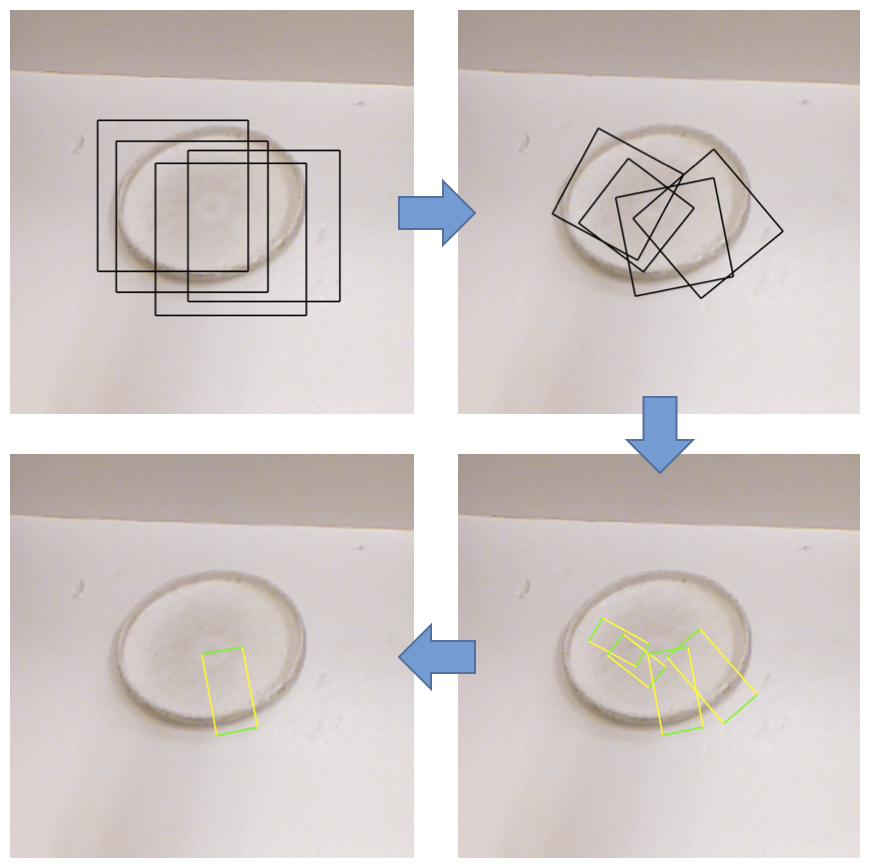}
      \caption{Another result of our proposed grasp detection method for a difficult object to detect grasp configuration. 
      Top left figure is the output of 
      the first STN (4 locations) and top right is the output of the second STN with rotation.
      Then, the bottom right is the output of the last STN with scaling and fine tuned location.
      The bottom left is the best graspable configuration among 4 outputs of the last STN.}
      \label{fig:result2}
      \vskip -0.2in
   \end{figure}
   
\subsection{Qualitative Results of Proposed Method}

   Fig.~\ref{fig:result1} shows an example of the step-by-step result using our proposed classification based grasp detection
   method using multiple-stage STN. The first figure illustrates detected 4 candidate locations from the first STN,
   then rotated candidates from the second STN for each output of the first STN, scaled and location
   adjusted candidates from the third STN, and finally chosen one best graspable rectangle from the grasp classifier.
   Fig.~\ref{fig:result2} illustrates another example of the result from our proposed method for an object that is difficult 
   to detect robotic grasp configuration with a single output regression model~\cite{Redmon:2015eq}. 
   Thanks to a number of candidates that were generated from the first STN, 
   this case also successfully identify robotic grasp configuration. Note that intermediate steps are observable partially
   (for 4 candidates in this case) so that it is easier to identify problems for the network than black-box models.
   Note that this partially observable feature was helpful to design and train our proposed network properly.
   
\subsection{Comparison Results for Proposed Method}

    \begin{figure}[!b]
      \centering
      \includegraphics[width=0.75\linewidth]{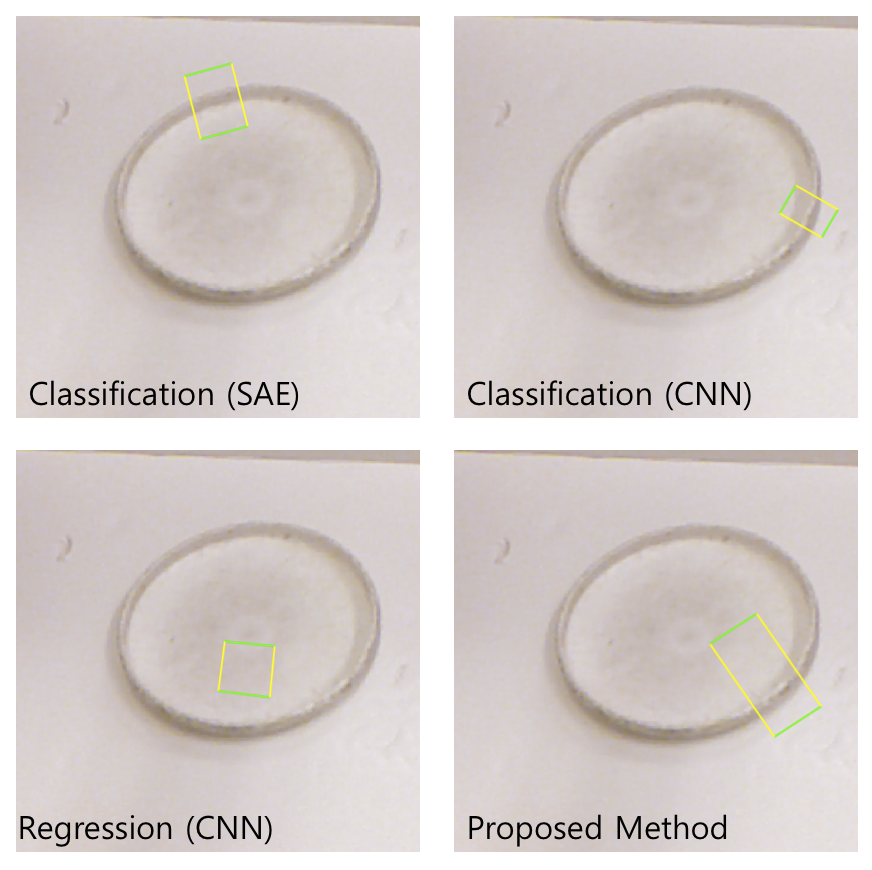}
      \vskip -0.1in
      \caption{Four comparison results from classification based grasp detection using sliding window with SAE and CNN,
      regression based grasp detection and the proposed classification based grasp detection using multi-stage STN.}
      \label{fig:result3}
   \end{figure}

    \begin{figure}[!b]
      \centering
        \vskip -0.1in
      \includegraphics[width=0.75\linewidth]{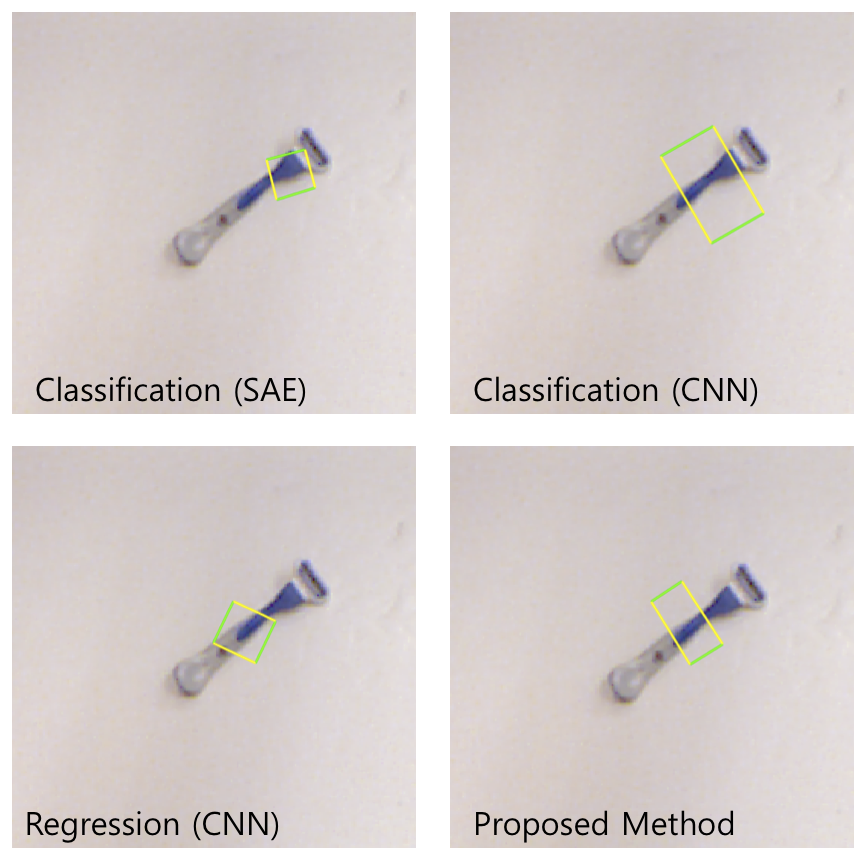}
      \vskip -0.1in
      \caption{Four comparison results from classification based grasp detection using sliding window with SAE and CNN,
      regression based grasp detection and the proposed classification based grasp detection using multi-stage STN.}
      \label{fig:result4}
   \end{figure}

Fig.~\ref{fig:result3} illustrates the grasp configuration outputs of classification (SAE),  classification (CNN),
regression (CNN), and proposed multiple stage STN based grasp detection methods.
As also reported in~\cite{Redmon:2015eq}, regression based grasp detection method yielded average of all good candidates,
while classification based methods yielded good grasp configurations.
Fig.~\ref{fig:result4} shows that classification (SAE) and regression based methods yielded relatively poor grasp configurations,
while our proposed method yielded good grasp configurations. Table~\ref{table_example} shows that
this superior performance of our proposed method is not just for a few images, but for all test images in general.
Our method achieved state-of-the-art performance with real-time processing speed.

\begin{table}[!t]
\caption{Performance for different methods in accuracy \& speed.
All measured on the same platform with the same size input.}
\vskip -0.1in
\label{table_example}
\begin{center}
\begin{tabular}{|c||c|c|}
\hline
Method & Accuracy (\%) & Time / Image \\
\hline
\hline
Classification (SAE) & 76.00 & 13 sec \\
\hline
Classification (CNN)  & 82.53 & 13 sec \\
\hline
Regression (CNN) & 70.67 & 11.3 msec \\
\hline
Our Single Stage STN & 71.30 & 13.6 msec \\
\hline
Our Multiple Stage STN & 89.60 & 23.0 msec \\
\hline
\end{tabular}
\end{center}
\vskip -0.3in
\end{table}

\section{DISCUSSION AND CONCLUSION}

In this paper, we proposed a novel classification based robotic grasp detection method using our multiple stage STN
and demonstrated that our proposed method achieved state-of-the-art performance in accuracy and real-time
computation time for relatively high resolution images. Our proposed method also has merits such as
easy integration with robot control algorithm for end-to-end training,
partially observable intermediate steps, 
and easy training without requiring pre-training with massive amount of data such as ImageNet.

In our results, the accuracy for regression based grasp detection method using ResNet-32 can be further improved
when using pre-training. However, in many massive image dataset such as ImageNet, depth images are not 
available so that it makes challenging to pre-train a model with multimodal information.
Regression based grasp detection methods have merit in terms of computation speed,
while our proposed method is relatively easy to train without pre-training.

\addtolength{\textheight}{-12cm}   









\bibliographystyle{IEEEtran}
\bibliography{grasp}

\end{document}